\documentclass[12pt]{l4dc2020} 

\usepackage{graphicx}      % include this line if your document contains figures

\newcommand{\tp}{{\mkern-1.5mu\mathsf{T}}}

\newcommand{\version}[2]{#1} % for arXiv version

\title[On Simulation and Trajectory Prediction with Gaussian Process Dynamics]{On Simulation and Trajectory Prediction\\ with Gaussian Process Dynamics}

  \author{%
\Name{Lukas Hewing}$^\text{1}$ \Email{lhewing@ethz.ch}\\
\Name{Elena Arcari}$^\text{1}$ \Email{earcari@ethz.ch}\\
\Name{Lukas P. Fr\"{o}hlich}$^\text{1,2}$ \Email{lukasfro@ethz.ch}\\
\Name{Melanie N. Zeilinger}$^\text{1}$ \Email{mzeilinger@ethz.ch}\\
\addr $^\text{1}$Institute for Dynamic Systems and Control, ETH Zurich, Zurich, Switzerland \\
\addr $^\text{2}$Bosch Center for Artificial Intelligence, Renningen, Germany}

\begin{document}
\maketitle
%%%%%%%%%%%%%%%%%%%%%%%%%%%%%%%%%%%%%%%%%%%%%%%%%%%%%%%%%%%%%%%%%%%%%%%%%%%%%%%%%%%%%%%%%%%%%%%%%%%%
%
%===================================================================================================
\begin{abstract} 
%===================================================================================================%    
Established techniques for simulation and prediction with Gaussian process (GP) dynamics often
implicitly make use of an independence assumption on successive function evaluations of the dynamics
model. This can result in significant error and underestimation of the prediction uncertainty,
potentially leading to failures in safety-critical applications. This paper discusses methods that
explicitly take the correlation of successive function evaluations into account. We first describe
two sampling-based techniques; one approach provides samples of the true trajectory distribution,
suitable for `ground truth' simulations, while the other draws function samples from basis function
approximations of the GP. Second, we propose a linearization-based technique that directly provides
approximations of the trajectory distribution, taking correlations explicitly into account. We
demonstrate the procedures in simple numerical examples, contrasting the results with established
methods.
\end{abstract}

%===================================================================================================
\section{Introduction}
%===================================================================================================
%
Gaussian process (GP) regression has become a popular tool for learning dynamic systems from data,
since it is flexible, requires little prior process knowledge and inherently provides a measure of
model uncertainty by providing a probabilistic distribution over function values. Various use-cases
have been proposed in the literature, such as state estimation~(\cite{Ko2008, Deisenroth2009a}),
model-based reinforcement learning~(\cite{Kuss2004, Deisenroth2009, Deisenroth2015}) and model
predictive control~(\cite{Klenske2013, Kocijan2016, Ostafew2016, Kamthe2018, Hewing2018a,
Kabzan2019}), see also~\cite{Hewing2020}.

All these problems require predictions of state $x$ under input $u$ for dynamic systems
\begin{equation} \label{eq:GP_dynModel}
    x_{k+1} = f(x_k,u_k) + w_k,
\end{equation}
where $w_k$ are i.i.d.\ Gaussian disturbances and the uncertain dynamics function is distributed
according to a Gaussian process $f\sim \mathcal{GP}$. The resulting stochastic state distributions
over a prediction horizon typically need to be numerically approximated, since no closed-form
solution exists. Established efforts are based on successive approximate and independent evaluations
of $f$ from uncertain inputs at each time step, such that the approximate distribution of the state
trajectory can be iteratively computed~(\cite{Girard2003, Pan2017}). Such an approach implicitly
introduces an independence assumption for successive evaluations of the function $f$, and neglects
the fact that GPs describe a distribution over functions, such that successive evaluations of $f$
are typically highly correlated. Figure~\ref{fg:MotExample} \emph{(top)} shows illustrations of
function samples from different GPs. Simulations with GP models avoiding this assumption have
recently been discussed in \cite{Umlauft2018, Bradford2019} and are used for
inference in GP state space models~(\cite{Ialongo2019}). In many previous results, however, both
theoretical analyses (e.g.\ in~\cite{Vinogradska2016,Beckers2016, Polymenakos2019}) and simulations
(e.g.\ in~\cite{Girard2003}) are carried out under the independence assumption. 

The goal of this paper is to provide simulation and prediction methods for
system~\eqref{eq:GP_dynModel} which take correlations of successive GP function evaluations into
account. Since the nonparametric stochastic description of the dynamics in~\eqref{eq:GP_dynModel} as
a GP can be intuitively less accessible and challenging to simulate, we begin by providing an
interpretation of the independence assumption in a motivating example using an analogy to parametric stochastic systems. The assumption then corresponds to considering all
uncertainty as process noise, rather than constant uncertain model parameters. We address the
correlation in successive function evaluations in~\eqref{eq:GP_dynModel} with methods for
sampling-based simulation and for direct approximation of the predicted state trajectory
distribution. We first discuss two sampling-based procedures: one
provides trajectory samples over the prediction horizon from the true trajectory distribution,
similarly proposed e.g.\ by \cite{Umlauft2018}, for which we provide an alternative view in terms of
the joint distributions of the trajectory, highlighting the specific computational structure and
complexity. We contrast this ground-truth simulation against a second simulation approach based on approximate function samples, enabled by basis function GP approximation
methods, similarly utilized in~\cite{Bradford2018} in the context of Bayesian optimization. For
direct approximation of the trajectory distribution, we then propose a modification of established
linearization-based uncertainty propagation methods (\cite{Girard2003}), which takes the correlation
of successive function evaluations into account. The procedures are demonstrated by revisiting the
motivating example.

\version{Appendix~\ref{app:GP_reg}}{Appendix A in~\cite{Hewing2019}} provides relevant background material on vector-valued GP regression and a more detailed definition of some of the notation used, which we believe to be intuitively accessible to readers familiar with GP regression.
%
%===================================================================================================
\section{Motivating Example}\label{subsec:MotExample}
%===================================================================================================
%
%
 \begin{figure}
    \mbox{
    \renewcommand\thesubfigure{1a}
    %\subfigure[Constant Offset]{
    \subfigure[Constant offset]{
        \begin{minipage}{.27\textwidth}
            \includegraphics{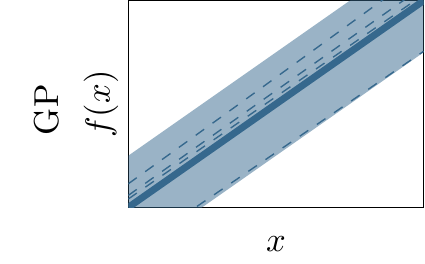} \\
            \includegraphics{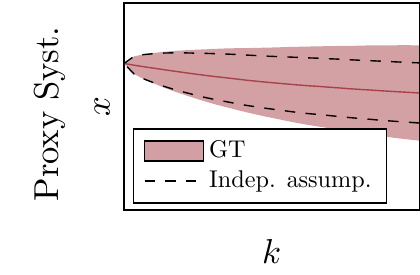}
        \end{minipage}}
    \renewcommand\thesubfigure{1b}        
    \subfigure[Additive noise]{
        \begin{minipage}{.205\textwidth}
            \includegraphics{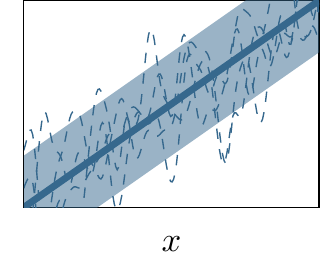} \\ 
            \includegraphics{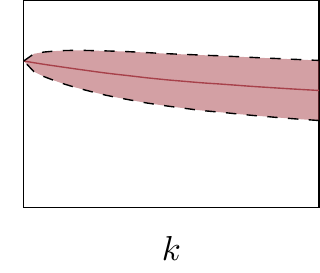}    
        \end{minipage}}
    \renewcommand\thesubfigure{2a}                
    \subfigure[\mbox{Uncertain gain}]{
        \begin{minipage}{.20\textwidth}
            \includegraphics{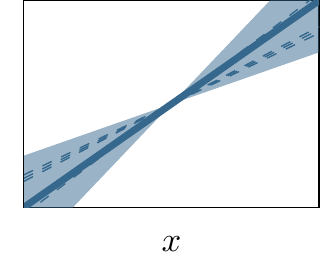} \\ 
            \includegraphics{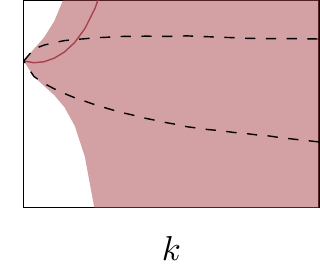}    
        \end{minipage}}        
    \hspace{-0.795\textwidth}   
    \renewcommand\thesubfigure{2b}        
    \subfigure[Mult.\ noise]{
        %\begin{minipage}{.21\textwidth}
            \begin{minipage}{\textwidth}
                \begin{flushright}
            \includegraphics{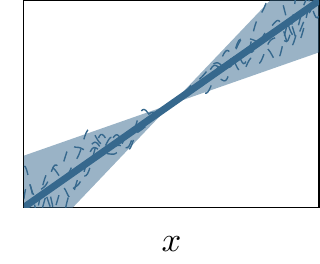} \\ 
            \includegraphics{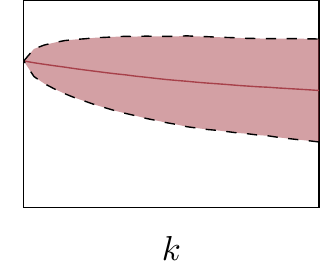}
            \end{flushright}
        \end{minipage}
        \hspace{0.8\textwidth}}} \vspace{-0.5cm}
 \caption{\emph{(top)}: Different GPs with mean, $2$-$\sigma$ variance and some function samples; the pairs \emph{(1a)/(1b)} and \emph{(2a)/(2b)} lead to the same mean and variance, respectively, when evaluated at any $x$ independently. \emph{(bottom)}: Mean and $2$-$\sigma$ variance of 20\,000 simulated trajectories from proxy systems of the above GPs. Dashed lines illustrate the variance from simulations under the independence assumption, which leads to identical results for \emph{(1a)} and \emph{(1b)} as well as \emph{(2a)} and \emph{(2b)}.} \label{fg:MotExample} \vspace{-0.5cm}
 \end{figure}
Since the accurate simulation of GP dynamics~\eqref{eq:GP_dynModel} is not
straightforward (see Section~\ref{subsec:sampling}), we begin by illustrating the implications of
the independence assumption through simple parametric proxy systems, for which trajectory samples can be easily obtained. As we will show by revisiting the
example in Section~\ref{sec:exampleRevisited}, these systems can be understood as limit cases of
certain GP dynamics with either increasing or vanishing kernel length scale,
illustrated in Figure~\ref{fg:MotExample} \emph{(top)}. We consider scalar autonomous systems
subject to i.i.d.\ noise $w_k \sim \mathcal{N}(0,1)$ and either
an uncertain element $\theta \sim \mathcal{N}(0,\sigma_f^2)$ which is constant for all
time steps, or 
a changing $\theta_k \sim \mathcal{N}(0,\sigma_f^2)$ independently and identically distributed.
The considered proxy systems are
\vspace{-0.07cm}
    \begin{enumerate}   \setlength\itemsep{0em}
        \item[\emph{(1a)}] \emph{Constant offset:} $x_{k+1} = 0.95 x_k + \theta + w_k$. This is a
        proxy for a GP prior with mean function $\mu(x) = 0.95x$ and squared exponential (SE) kernel
        $k_{\text{SE}}(x,x') = \sigma_f^2 \exp (-\frac{(x-x')^2}{2 \ell^2})$ with large length scale
        $\ell = 10$.
    \item[\emph{(1b)}] \emph{Additive noise:} $x_{k+1} = 0.95 x_k + \theta_k + w_k$. This is a proxy for a GP prior with mean function $\mu(x) = 0.95x$ and SE kernel with small length scale $\ell = 0.1$.
        \item[\emph{(2a)}] \emph{Uncertain gain:} $x_{k+1} = (0.95 + \theta) x_k + w_k$. This is a
    proxy for a GP prior with mean function $\mu(x) = 0.95x$ and a linear kernel
    $k_{\text{lin}}(x,x') = \sigma_f^2 x x'$.
        \item[\emph{(2b)}] \emph{Multiplicative noise:} $x_{k+1} = (0.95 + \theta_k) x_k$. This is a proxy for a GP prior with mean function $\mu(x) = 0.95x$ and multiplied linear/SE kernel with small length scale $\ell=0.1$.
    \end{enumerate}
%\end{description}
The plots in Figure~\ref{fg:MotExample} \emph{(bottom)} show mean and variance
information from 20\,000 simulated state trajectories of these proxy systems.
The independent uncertainty $\theta_k$ at each time step in \emph{(1b)} and \emph{(2b)}, in contrast to
random but constant $\theta$ in \emph{(1a)} and \emph{(2a)}, leads to a significantly smaller spread
of system trajectories, i.e.\ significantly smaller `predicted' uncertainty as indicated by the
variance in Figure~\ref{fg:MotExample}. Often times, methods for prediction and simulation with GP
dynamics consider successive evaluations of the dynamics $f$ as independent. This corresponds to
drawing the persistent uncertain parameter $\theta$ in \emph{(1a)} and \emph{(2a)} independently at
each time step, i.e.\ identical to $\theta_k$ in \emph{(1b)} and \emph{(2b)}, which leads to a
significant underestimation of the resulting uncertainty, shown with dashed lines in
Figure~\ref{fg:MotExample} \emph{(bottom)}.
%
%===================================================================================================
\section{Predictions with GP Dynamics}\label{sec:GP_predictions}
%===================================================================================================
%
In this paper, we provide simulation and prediction methods for system~\eqref{eq:GP_dynModel} which take correlations of successive GP function evaluations into
account and enable accurate predictions in examples \emph{(1a)} and \emph{(2a)}.
For simplicity, we focus on
autonomous systems
\begin{align} \label{eq:GP_dynModel_aut}
    x_{k+1} = f(x_k) + w_k, \ f \sim \mathcal{GP}(\mu, k),
\end{align}
subject to i.i.d.\ noise $w_k \sim \mathcal{N}(0, Q)$. The distribution of $f$ is specified by a GP 
with mean function $\mu : \mathbb{R}^{n} \rightarrow \mathbb{R}^{n}$, which in vector-valued 
regression maps to a vector, and the kernel $k : \mathbb{R}^{n} \times \mathbb{R}^{n} \rightarrow \mathbb{R}^{n \times n}$, 
which maps to a positive (semi-)definite variance matrix. All methods can be applied similarly 
to controlled systems, see \version{Appendix~\ref{app:controlled_sys}}{Appendix C in~\cite{Hewing2019}}.

In the following, we use respective capital letters for stacked vectors, e.g.\ $X = [x^\tp_1,
\ldots, x^\tp_N]^\tp$ and $X_{a:b}$ with indices $a < b$ to refer to $[x^\tp_a, \ldots,
x^\tp_b]^\tp$. With a slight abuse of notation we overload $\mu$ and $k$ when evaluated on stacked
vectors, such that $\mu(X)_i = \mu(x_i)$ and $k(X,X')_{i,j} = k(x_i,x'_j)$. With this, we write
evaluations of $f$ at $x_a,\ldots, x_b$ compactly as $[f^\tp_a,\ldots,f^\tp_b]^\tp \sim
\mathcal{N}(\mu(X_{a:b}),k(X_{a:b},X'_{a:b}))$. 
\begin{remark}[Inference] \label{rm:inference}
Inference given data $\mathcal{D} = \{ X^\text{t}, Y^\text{t}\}$ is carried out by conditioning the Gaussian distributions on the collected data points. This can be expressed as modified mean and kernel function
\begin{align*}
    \mu^{\mathcal{D}}(x) &= \mu(x) + k(x,X^\text{t})(k(X^\text{t},X^\text{t}) + I \otimes Q)^{-1} (Y^\text{t} - \mu(X^t)), \\
    k^{\mathcal{D}}(x,x') &= k(x,x') - k(x,X^\text{t})(k(X^\text{t},X^\text{t}) + I \otimes Q)^{-1}k(X^\text{t},x').
\end{align*}
To simplify notation, we therefore refer simply to~$\mu$ and $k$ for the remainder of the paper, which can similarly denote a GP conditioned on data.
\end{remark}

Under the GP assumption, all evaluations of $f$ are jointly Gaussian distributed according to the specified mean and kernel function. Using~\eqref{eq:GP_dynModel_aut}, we can therefore express the distribution of the predicted state trajectory over $N$ time steps implicitly as
\begin{align}
    &\begin{bmatrix}
        x_1 \\ \vdots \\ x_N
    \end{bmatrix} \sim \mathcal{N} \!\left(\!
    \begin{bmatrix}
        \mu(x_0) \\ \vdots \\ \mu(x_{N\!-\!1})
    \end{bmatrix}\!, \! 
        \begin{bmatrix} k(x_0,x_0) + Q & \ldots & k(x_0,x_{N\!-\!1}) \\ 
                        \vdots & \ddots & \vdots \\
                        k(x_{N\!-\!1},x_0) & \ldots & k(x_{N\!-\!1},x_{N\!-\!1}) + Q
        \end{bmatrix} \! \right) \label{eq:implicitDist},
\end{align}
which, using shorthand notation, can be compactly expressed as
\[ 
 X_{1:N} \sim \mathcal{N}(\mu(X_{0:N-1}),k(X_{0:N-1},X_{0:N-1}) + I \otimes Q).
\]
This implicit description of the predicted trajectory is a challenging object to deal with, since
the (shifted) state sequence appears on both the left- and right-hand side. 
\begin{remark}[Independence Assumption]\label{rm:indAssumption}
    The common independence assumption can be understood as approximating~\eqref{eq:implicitDist} using a block diagonal covariance matrix
    \begin{align*}
        &\begin{bmatrix}
            x_1 \\ \vdots \\ x_N
        \end{bmatrix} \sim \mathcal{N} \!\left(\!
        \begin{bmatrix}
            \mu(x_0) \\ \vdots \\ \mu(x_{N\!-\!1})
        \end{bmatrix}\!, \! 
            \begin{bmatrix} k(x_0,x_0) + Q & \ldots & 0 \\ 
                            \vdots & \ddots & \vdots \\
                            0 & \ldots & k(x_{N\!-\!1},x_{N\!-\!1}) + Q
            \end{bmatrix} \! \right).
    \end{align*}
\end{remark}
In the following, we provide sampling-based methodologies that avoid the independence assumption,
allowing for accurate simulation of the dynamic system, as well as a linearization-based direct
approximation of the trajectory distribution~\eqref{eq:implicitDist}.
%
%---------------------------------------------------------------------------------------------------
\subsection{Sampling-Based Simulation with GP Dynamics}\label{subsec:sampling}
%---------------------------------------------------------------------------------------------------
%
Sampling-based simulation of systems given by GP dynamics is of great interest, e.g.\ for particle
filtering or simulation-based or -aided controller and system design. A naive approach is to draw
$N_s$ \emph{function samples} $f^{(i)} \sim \mathcal{GP}(\mu,k)$, $i = 1,\ldots,N_s$, as shown in
Figure~\ref{fg:MotExample}. These function samples can then be used in simulation to generate state
trajectory samples $X^{(i)} = [(x_0^{(i)})^\tp, \ldots, (x_N^{(i)})^\tp]^\tp$.
Generally, function samples can only be obtained approximately, which is
typically done by
evaluating $f$ on a fine grid spanning the entire domain and subsequent sampling from the resulting
joint Gaussian distribution. It therefore comes with the usual limitations associated with gridding,
such as very poor scalability to systems of higher dimensions.

Here, we discuss two sampling-based approaches alleviating this drawback. The first approach
directly generates samples of the \emph{trajectory} from the true trajectory distribution,
circumventing the need to draw a function sample. The second approach generates \emph{approximate}
function samples based on basis function approximations of the GP. This avoids gridding by providing
explicit representations of the sampled functions and can have computational advantages over the
direct trajectory sampling.

\subsubsection{Trajectory Sampling}\label{subsubsec:TrajSampling}
Given the initial state $x_0$, we generate trajectory samples $X^{(i)}$ which are
consistent with the GP~\eqref{eq:implicitDist} by drawing samples of subsequent time steps, starting
with the first step
\[ x^{(i)}_1 = \mu(x_0) + \sqrt{k(x_0,x_0) + Q} \; \tilde{w}_0^{(i)}, \] where $\sqrt{\cdot}$
denotes the Cholesky decomposition and $\tilde{w}_0^{(i)} \sim \mathcal{N}(0,I)$ is drawn from the
standard normal distribution. Samples of the following state $x^{(i)}_2$ can then be drawn by
conditioning the GP on the respective realization, i.e.\ $\mathcal{D}^{(i)}_1 = \{x_1^{(i)},
x^{(i)}_0\}$ in Remark~\ref{rm:inference}, see also~\cite{Umlauft2018}. Equivalently, one can consider the joint distribution of
$x^{(i)}_1$ and $x^{(i)}_2$, avoiding the computation of the conditional distributions and revealing
the particular structure of the problem
\begin{align*} 
&\begin{bmatrix} x^{(i)}_1 \\ x^{(i)}_2 \end{bmatrix} = \begin{bmatrix} \mu(x_0) \\ \mu(x^{(i)}_1) \end{bmatrix} + \sqrt{\begin{bmatrix} k(x_0,x_0) + Q & k(x_0,x^{(i)}_1) \\ k(x^{(i)}_1,x_0) & k(x^{(i)}_1,x^{(i)}_1) + Q \end{bmatrix}} \begin{bmatrix} w^{(i)}_0 \\ w^{(i)}_1 \end{bmatrix}.
\end{align*}
Given the sample $x^{(i)}_1$ of the first time step, the Cholesky decomposition for the joint
distribution can be computed, and $x^{(i)}_2$ generated using samples of the standard normal
distribution $w_1^{(i)} \sim \mathcal{N}(0,I)$. Iterating this procedure, a sample up to time step
$k+1$ is generated by
\begin{align}\label{eq:SamplingIteration}
X_{1:k+1}^{(i)} = \mu(X^{(i)}_{0:k}) + \sqrt{k(X^{(i)}_{0:k},X^{(i)}_{0:k}) + I \otimes Q} \, \tilde{W}^{(i)}_{0:k+1},
\end{align}
where $X^{(i)}$ is the trajectory sample, and $\tilde{W}^{(i)} =
[(\tilde{w}^{(i)}_0)^\tp,\ldots,(\tilde{w}^{(i)}_{N-1})^\tp]^\tp$ the corresponding sample drawn
from the standard normal distribution. 
Since there is no approximation involved, this procedure can be used for \emph{ground truth}
simulations and serves as a basis for comparison against other methods in the following. Note that
it is not necessary to recalculate the entire Cholesky decomposition at each time step, but only the
added lines of each time step, cf.\ the standard algorithm for Cholesky
decomposition~\citep[Ch.~2.7]{Demmel1997}. The computational complexity is therefore related to
\emph{one} full Cholesky decomposition and scales cubically with the prediction horizon. 
%
%---------------------------------------------------------------------------------------------------
\subsubsection{Approximate Function Samples Based on Basis Functions}\label{sec:approximate_function_samples}
%---------------------------------------------------------------------------------------------------
%
The computational cost of drawing trajectory samples from the true distribution grows rapidly with
the prediction horizon, while generating function samples based on gridding scales poorly with the
state dimension $n$. This motivates an alternative procedure of generating explicit representations
of \emph{approximate} function samples to be used for simulation--resulting in computational
complexity that is linear in the prediction length while scaling to higher dimensional systems. The
approach is based on the fact that GPs can be written as a linear combination of a (possibly
infinite) number of basis functions
\begin{align*}
 f(x) = \phi(x)^\tp \theta  \text{, with } \theta \sim \mathcal{N}(\mu_\theta, \Sigma_\theta),
\end{align*}
where the kernel is given by $k(x, x') = \phi(x)^\tp  \phi(x')$ and $\mu_\theta, \Sigma_\theta$
follow from Bayesian linear regression, see e.g.~\citet[Ch.~2.1]{Rasmussen2006Book}. The function
$f$ can then be approximated by a \emph{finite} subset of $m$ basis functions, and an explicit
approximate function sample $\tilde{f}^{(i)} = \sum_{j=1}^{m} \phi_j(x) \theta_j^{(i)}$ can be
constructed by sampling the weight vector $\theta$.

In general, computing basis functions $\phi(x)$ for a particular kernel can be challenging. A
common finite dimensional approximation of $\phi$ is given by the first $m$ kernel eigenfunctions,
which can be approximated by the Nystr\"{o}m method if the solutions are not analytically
tractable~\citep[Ch.~19.1]{Press2007Book}. For continuous shift-invariant kernels, \cite{Rahimi2008}
have shown that cosine functions with random frequencies and phase shifts are particularly
well-suited as basis functions, leading to the so-called sparse spectrum GP approximation
\citep{Lazaro2010}, the use of which was proposed in~\cite{Bradford2018} to draw approximate
function samples in the context of Bayesian optimization. Typically, the kernel approximations
recover the exact distribution of $f$ as the number of basis functions approaches infinity.
\version{Appendix~\ref{app:basis_func} includes}{Appendix B in~\cite{Hewing2019} includes} a short
discussion of common basis function approximations as used in our numerical examples in
Section~\ref{sec:exampleRevisited}.
%
%---------------------------------------------------------------------------------------------------
\subsection{Linearization-Based Approximation}\label{subsec:linearization}
%---------------------------------------------------------------------------------------------------
%
In many applications, such as predictive control, the computational cost of sampling-based
GP predictions can be prohibitive, motivating efficient direct approximations of the predictive
trajectory distributions. In the following, we propose a linearization-based approximation method
which can be understood as a modification of established prediction techniques. The procedure is
based on approximating the distribution of the predicted state sequence by iteratively
linearizing the mean function $\mu$ around the previous state mean. We follow a similar procedure as
in Section~\ref{subsubsec:TrajSampling} to obtain
\begin{equation*}\label{eq:firstStepDistribution}
 x_1 = \mu(x_0) + \sqrt{k(x_0,x_0) + Q} \, \tilde{w}_0
\end{equation*}
in which again $\tilde{w}_0 \sim \mathcal{N}(0,I)$. Similar to~\cite{Girard2003}, we consider a first-order Taylor expansion of $\mu$ around $\mu_1 = \mu(x_0)$ for the next time step, but consider the zero-order expansion of the \emph{joint} variance $k$. This leads to
\begin{align*} 
\begin{bmatrix} x_1 \\ x_2 \end{bmatrix} &\approx \begin{bmatrix} \mu_1 \\ \mu(\mu_1) \end{bmatrix} \nonumber + \begin{bmatrix} I & 0 \\ \nabla\mu(\mu_1) & I \end{bmatrix} \sqrt{\begin{bmatrix} k(x_0,x_0) + Q & k(x_0,\mu_1) \\ k(\mu_1,x_0) & k(\mu_1,\mu_1) + Q \end{bmatrix}} \begin{bmatrix} \tilde{w}_0 \\ \tilde{w}_1 \end{bmatrix}.
\end{align*}
Iterating this procedure and introducing the short-hand notation $\tilde{W} =
[\tilde{w}^\tp_0,\ldots,\tilde{w}^\tp_{N-1}]^\tp$ and $M = [\mu_0^\tp, \ldots, \mu_N^\tp]^\tp$ in
which $\mu_0 = x_0$ and $\mu_{k+1} = \mu(\mu_k)$, yields the expression 
\[ 
 X_{1:N} \approx M_{1:N} + A \sqrt{k(M_{0:N-1},M_{0:N-1}) \! + \! I \otimes Q} \, \tilde{W},
\] 
in which $A$ is a lower triangular block matrix with the relevant derivatives $A_{i,j} =
\prod_{l=j}^{i-1}\nabla \mu(\mu_l)$. This results in the following approximate distribution of the
state trajectory
\begin{align}\label{eq:linApproximataionDist}
    X_{1:N} \dot \sim \mathcal{N}(M_{1:N}, A (k(M_{0:N-1},M_{0:N-1}) \! + \! I \otimes Q) A^\tp).
\end{align}
In contrast to established prediction methods using the independence assumption, we approximate the
full variance matrix in~\eqref{eq:implicitDist} instead of assuming a block diagonal structure, see
Remark~\ref{rm:indAssumption}. The proposed technique therefore corrects for correlation of
subsequent function evaluations along the mean prediction. Note that no Cholesky decomposition is
necessary for~\eqref{eq:linApproximataionDist} or computation of the mean trajectory $M_{1:N}$. The
computational complexity therefore scales quadratically with the prediction horizon, as opposed to a
linear scaling under the independence assumption with a block diagonal structure. 
\begin{figure}
    \mbox{\renewcommand\thesubfigure{1a}
    \subfigure[SE, $\ell = 10$]{
        \begin{minipage}{.27\textwidth}
            \includegraphics{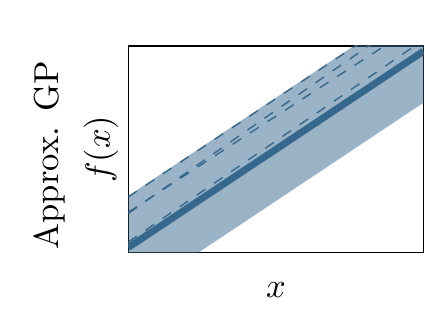} \\ 
            \includegraphics{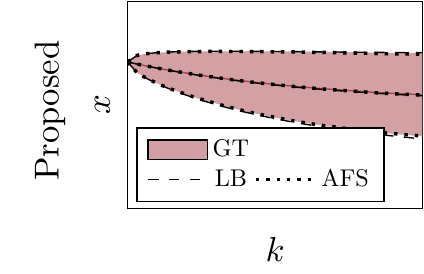}
        \end{minipage}}
    \renewcommand\thesubfigure{1b}        
    \subfigure[SE, $\ell = 0.1$]{
        \begin{minipage}{.2\textwidth}
            \includegraphics{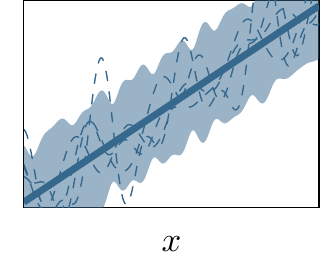} \\ 
            \includegraphics{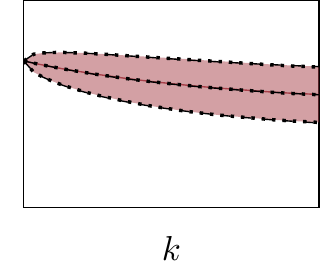}    
        \end{minipage}}
    \renewcommand\thesubfigure{2a}                
    \subfigure[Linear]{
        \begin{minipage}{.2\textwidth}
            \includegraphics{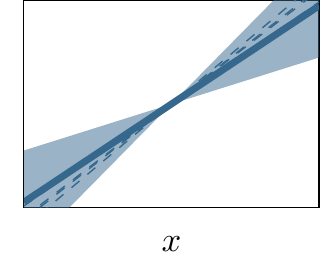} \\ 
            \includegraphics{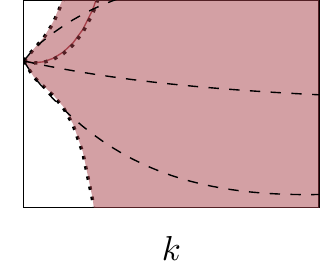}    
        \end{minipage}}        
    \hspace{-0.795\textwidth}   
    \renewcommand\thesubfigure{2b}        
    \subfigure[Mult., $\ell \!=\! 0.1$]{
        \begin{minipage}{\textwidth}
            \begin{flushright}
            \includegraphics{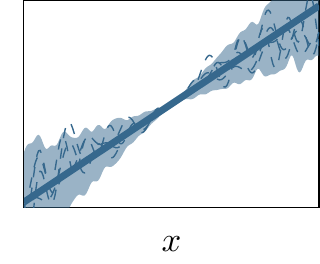} \\ 
            \includegraphics{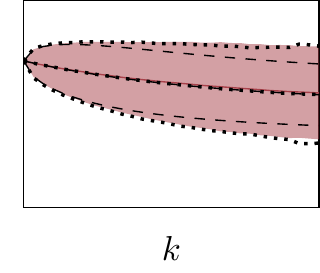}    
        \end{flushright}
    \end{minipage}\hspace{0.8\textwidth}}}
\caption{\emph{(top):} Approximate GPs using 10 basis functions. \emph{(bottom):} Proposed prediction techniques. Mean and $2$-$\sigma$ variance of 20\,000 ground truth (GT) simulations using trajectory sampling in red. The results using approximate function sampling (AFS) as dotted lines and linearization-based (LB) prediction as dashed lines.} \label{fg:ExampleRev}
%\end{figure*}
\end{figure}
%
%===================================================================================================
\section{Example Revisited}\label{sec:exampleRevisited}
%===================================================================================================
%
In order to demonstrate the proposed simulation and prediction methods, we revisit the motivating
example of Section~\ref{subsec:MotExample}. We make use of these simple examples, i.e.\ GPs without
explicitly considering data, to clearly illustrate the effects of different techniques and
assumptions. The techniques developed, however, apply similarly to GPs conditioned on data (cf.\
Remark~\ref{rm:inference}) or controlled systems (see
\version{Appendix~\ref{app:controlled_sys}}{Appendix C in~\cite{Hewing2019}}). The trajectory
sampling technique described in Section~\ref{subsubsec:TrajSampling} allows us to simulate the
actual GP dynamics described in Section~\ref{subsec:MotExample} and shown in
Figure~\ref{fg:MotExample} \emph{(top)}, instead of the proxy systems.

We compare \emph{ground truth} simulations using trajectory sampling (GT) to simulations using
approximate function samples (AFS) and the developed linearization-based prediction technique (LB).
Approximate function samples are drawn using 10 random basis functions of the SE kernels for
examples \emph{(1a)} and \emph{(1b)}, which are transformed with a linear gain for \emph{(2b)}. For
\emph{(2a)} we use the exact basis function representation by the single linear eigenfunction.
Additional information can be found in \version{Appendix~\ref{app:basis_func}}{Appendix B
in~\cite{Hewing2019}}. The resulting basis function approximations are shown in Figure~\ref{fg:ExampleRev}
\emph{(top)}. Note that the approximation with as little as 10 basis functions is possible due to
the simple structure of the example. For complex applications, good approximations
can typically be found using hundreds to a few thousand basis functions~\citep{Lazaro2010,
Lobato2014}.

The results with the techniques considered are shown in Figure~\ref{fg:ExampleRev} \emph{(bottom)}.
Ground truth simulations using the trajectory sampling in Section~\ref{subsubsec:TrajSampling} show
very good correspondence to the proxy systems investigated in Section~\ref{subsec:MotExample} (cf.\
Figure~\ref{fg:MotExample}). All the methods show a clear adjustment to the persistent uncertainties
in \emph{(1a)} and \emph{(2a)}. This is in contrast to established prediction methods, which
consider \emph{(1a)} and \emph{(2a)} identical to \emph{(1b)} and \emph{(2b)}, since the marginal
distributions for a \emph{single} evaluation are identical, as illustrated by the shaded regions in
Figure~\ref{fg:MotExample} \emph{(top)}. For examples \emph{(1a)} and \emph{(1b)}, the
linearization-based uncertainty propagation technique developed in
Section~\ref{subsec:linearization} is almost perfect. This is different in \emph{(2a)} and
\emph{(2b)}, where the linearization-based technique leads to significant errors due to the local
approximation around the mean. Note here in particular, that as the predicted mean approaches the
origin, the local approximation of $k$ approaches zero. Simulation based on basis function
approximations yields almost indistinguishable results from the ground truth simulations. This
suggests that good results can be obtained at significantly lower computational cost than ground
truth simulations based on approximate function samples, if a suitable basis function representation
can be found.
%
%===================================================================================================
\section{Conclusion} 
%===================================================================================================
%
We have shown that correlation between successive function evaluations in a GP dynamics system
significantly influences the resulting state trajectory distributions. Two sampling-based methods
and a proposed direct approximation of the trajectory distributions were compared and shown to take
these correlations into account. The methods significantly improve prediction accuracy over
established techniques based on an independence assumption, at the cost of increased computational
demand for long horizons. We showed that approximate function samples obtained from basis function
approximations of the GP can alleviate this drawback, allowing computationally efficient
sampling-based simulations.

\acks{The authors would like to acknowledge support from the Swiss National Science Foundation, grant no. PP00P2\_157601 / 1.}

\bibliography{GP-Prediction}

\version{
%%%%%%%%%%%%%%%%%%%%%%%%%%%%%%%%%%%%%%%%%%%%%%%%%%%%%%%%%%%%%%%%%%%%%%%%%%%%%%%%%%%%%%%%%%%%%%%%%%%%
\appendix
%%%%%%%%%%%%%%%%%%%%%%%%%%%%%%%%%%%%%%%%%%%%%%%%%%%%%%%%%%%%%%%%%%%%%%%%%%%%%%%%%%%%%%%%%%%%%%%%%%%%
%
%===================================================================================================
\section{Vector-valued Gaussian Process Regression}\label{app:GP_reg}
%===================================================================================================
%
GP regression is a non-parametric framework for nonlinear regression under the statistical model 
\begin{equation*} y = f(x) + w, \end{equation*}
in which the unknown function $f$ maps inputs $x$ to outputs $y$ under i.i.d.\ noise $w \sim \mathcal{N}(0,Q)$. Note that this corresponds to the autonomus dynamic system~\eqref{eq:GP_dynModel_aut}, where input and output dimensions of $f$ are identical, i.e.\ $f : \mathbb{R}^{n} \rightarrow \mathbb{R}^{n}$.
The GP assumption states that all function values of $f$ are jointly Gaussian distributed according to mean function $\mu$ and kernel function $k$
\begin{equation*}
     \begin{bmatrix}
        f_1 \\ \vdots \\ f_N
    \end{bmatrix} \sim \mathcal{N}\left( 
    \begin{bmatrix}
        \mu(x_1) \\ \vdots \\ \mu(x_N)
    \end{bmatrix}, 
        \begin{bmatrix} k(x_1,x_1) & \ldots & k(x_1,x_N) \\ 
                        \vdots & \ddots & \vdots \\
                        k(x_N,x_1) & \ldots & k(x_N,x_N)
        \end{bmatrix}
    \right), \label{eq:GP_jointDist} 
\end{equation*}
i.e.\ the distribution of function values $f_1, \ldots, f_N$ is parameterized by the input locations $x_1, \ldots, x_N$ and the respective mean and kernel functions $\mu$ and $k$ of the GP. This assumption on the function $f$ is therefore typically expressed as
\begin{equation*} 
    f \sim \mathcal{GP}(\mu, k), 
\end{equation*}
emphasizing that GPs describe a random distribution of the \emph{function}. Note here that in-vector
valued GP regression the mean function maps to a vector $\mu : \mathbb{R}^{n} \rightarrow
\mathbb{R}^{n}$ and the kernel to a positive (semi-)definite variance matrix $k : \mathbb{R}^{n}
\times \mathbb{R}^{n} \rightarrow \mathbb{R}^{n \times n}$.
The kernel function $k$ must be chosen such that the resulting variance 
is positive (semi-)definite. A number of choices of kernel functions $k$ are available, in
particular in the scalar setting $k^s: \mathbb{R}^{n} \times \mathbb{R}^{n} \rightarrow
\mathbb{R}$, for instance the squared exponential kernel
\[ k_{SE}^s(x,x') = \sigma_f^2 \exp\!\left(-\frac{1}{2\ell^2} \Vert x - x' \Vert^2\right), \] which
is parameterized by the length-scale $\ell$ and variance $\sigma_f^2$. Scalar-valued kernels can be
used to generate matrix valued kernel functions by assigning distance metric $d: \mathbb{N}
\rightarrow \mathbb{R}$ to each dimension $i = 1,\ldots,n$, in order to define a covariance between
different output dimension $[k(x,x')]_{i,j} = k_s([x^\tp, d(i)]^\tp, [{x'}^\tp, d(j)]^\tp)$, see
also~\cite{Alvarez2012}. The important case of independent output dimensions is given by 
\[ 
[k(x,x')]_{i,j} = \begin{cases} k^s(x,x'), &i = j \\ 0.  &\text{otherwise} \end{cases}
\]
This is equivalent to considering a separate GP for each output dimension independently, which is
often done in dynamics learning with GPs. 
%
%===================================================================================================
\section{Basis Function Approximation for Gaussian Processes}\label{app:basis_func}
%===================================================================================================
%
An alternative view of GP regression can be given as Bayesian linear regression with a potentially
infinite number of basis functions, see e.g.~\cite{Rasmussen2006Book}. Considering a finite number
of these basis functions therefore enables approximate explicit representations of $f \sim
\mathcal{GP}$ as $f = \phi(x)^\tp \theta$ with basis functions $\phi$ and weights $\theta \sim
\mathcal{N}(\mu_\theta,\Sigma_\theta)$. Given data, the distribution of $\theta$ is obtained from
Bayesian linear regression, while we can assume the prior distribution $\theta \sim
\mathcal{N}(0,I)$ without loss of generality, such that the resulting approximation of the kernel is
$k(x,x') \approx \phi(x)^\tp \phi(x')$. In the following, we present two approaches for computing
possible basis functions to be used in approximate GP regression, as discussed in
Section~\ref{sec:approximate_function_samples}.
%
%---------------------------------------------------------------------------------------------------
\paragraph{Eigenfunctions}
%---------------------------------------------------------------------------------------------------
%
Eigenfunctions $g(x)$ with corresponding eigenvalue $\lambda$ satisfy the relation
\begin{align*}
	\int k(x, x') g(x) p(x) dx = \lambda g(x')
\end{align*}
for a particular kernel $k(x, x')$ and density $p(x)$.
Informally, Mercer's Theorem \citep{Koenig2013Book} states that any kernel can be expressed in terms of its eigenbasis
\begin{align*}
	k(x, x') = \sum_{i=0}^{\infty} \lambda_i g_i(x) g_i(x').
\end{align*}
For non-degenerate kernels, an infinite number of eigenfunctions with non-zero eigenvalues exist.
For a finite approximation, the first $m$ eigenfunctions with largest eigenvalues are often used as
basis functions, such that
$\phi(x) = [\phi_1(x), \dots, \phi_{m}(x)]^\tp$ and $\phi_i(x) = g_i(x) / \sqrt{\lambda_i}$.
For some kernels, such as the squared exponential kernel and $p(x)$ the normal distribution, the
eigenfunctions can be computed analytically,
while in general, eigenfunctions can be approximated using e.g.\ the Nystr\"{o}m method.
We refer to
\citet[Ch.~4.3]{Rasmussen2006Book} for a more detailed discussion of the eigenfunctions of kernels.
%
%---------------------------------------------------------------------------------------------------
\paragraph{Random Fourier Features}
%---------------------------------------------------------------------------------------------------
%
Another popular choice of basis functions to approximate a (continuous and shift-invariant) kernel
is given by $g_i(x) = \cos(\omega_i x + b_i)$, where $\omega_i$ and $b_i$ are random variables
\citep{Rahimi2008}. In particular, $b_i \sim \mathcal{U}[0, 2\pi]$ and \mbox{$\omega_i \sim
p(\omega)$} where  $p(\omega)$ is proportional to the Fourier transform of the particular kernel at
hand. Bochner's theorem states that for any shift-invariant kernel, the Fourier dual is a proper
density function  \citep{Stein2012Book}. For the case where $k(x, x')$ is the SE kernel with
$\sigma_f^2 = 1$, $p(\omega)$ thus follows the normal distribution,
\begin{align*}
	p(\omega) \propto S(\omega) = \mathcal{F}\{ k^{SE}(r) \}  = (2\pi \ell )^{n/2} \exp\left( -2 \pi^2 \ell^2 \omega^2 \right),
\end{align*}
with $\ell$ being the kernel length scale and $r = |x - x'|$. Given an approximation with $m$ basis
functions, these are given by $\phi_i(x) = g_i(x)/\sqrt{m} $. We sample 10 random basis functions
together with corresponding weight vectors to generate our basis function approximation in
Section~\ref{sec:exampleRevisited}.
%
%===================================================================================================
\section{Controlled Systems}\label{app:controlled_sys}
%===================================================================================================
%
The methods developed can be similarly applied to controlled systems~\eqref{eq:GP_dynModel}, in which case we express
\[ f \sim \mathcal{GP}(\mu(z), k(z,z')),\]
by introducing the short hand notation $z = [x^\tp, u^\tp]^\tp$, such that $f$ is also a function of the input $u$.
%---------------------------------------------------------------------------------------------------
\paragraph{Trajectory sampling}
%---------------------------------------------------------------------------------------------------
The resulting iterative sampling procedure~\eqref{eq:SamplingIteration} then reads
\begin{align*}
    X_{1:k+1}^{(i)} = \mu(Z^{(i)}) + \sqrt{k(Z^{(i)}_{0:k},Z^{(i)}_{0:k}) + I \otimes Q} \, \tilde{W}^{(i)}_{0:k+1},
\end{align*}
in which $Z^{(i)} = [(x^{(i)}_0)^\tp,(u_0)^\tp,\ldots, (x^{(i)}_k)^\tp,(u_k)^\tp]^\tp$. Note that this formulation allows
deterministic gradient-based optimization, since only samples from the standard normal
distribution are required which can be drawn before optimization. This way, the procedure lends
itself e.g.\ to sampling-based model predictive control.
%---------------------------------------------------------------------------------------------------
\paragraph{Linearization-based Approximation}    
%---------------------------------------------------------------------------------------------------
For the linearization-based approximation, the approximate
distribution~\eqref{eq:linApproximataionDist} holds with modified $M = [\mu_0, \ldots, \mu_N ]$
and $\mu_0 = x_0$ and $\mu_{k+1} = \mu(\mu_k, u_k)$, and lower triangular block matrix $A$ with
blocks $ A_{i,j} = \prod_{l=j}^{i-1}\nabla_x \mu(\mu_l, u_l)$, 
where $\nabla_x$ is the partial gradient with respect to the state $x$.
}{}

\end{document}